\documentclass{article}
\usepackage[nonatbib,preprint]{nips_2018}

\usepackage{cite}

\usepackage[utf8]{inputenc} 
\usepackage[T1]{fontenc}    
\usepackage{hyperref}       
\usepackage{url}            
\usepackage{booktabs}       
\usepackage{amsfonts}       
\usepackage{nicefrac}       
\usepackage{microtype}      

\usepackage{graphicx}
\graphicspath{ {images/} }

\usepackage{algorithm}
\usepackage{algorithmic}

\def\vect#1{\mbox{\boldmath $#1$}}

\usepackage{wrapfig}

\title{C-SENN: Contrastive Self-Explaining\\ Neural Network}

\author{
  Yoshihide Sawada\\
  Tokyo Research Center, Aisin\\
  \texttt{yoshihide.sawada@aisin.co.jp} \\
  \And
  Keigo Namamura\\
  Tokyo Research Center, Aisin\\
  \texttt{keigo.nakamura@aisin.co.jp} \\
}

\begin{document}

\maketitle

\begin{abstract}
In this study, we use a self-explaining neural network (SENN), which learns unsupervised concepts, to acquire concepts that are easy for people to understand automatically. In concept learning, the hidden layer retains verbalizable features relevant to the output, which is crucial when adapting to real-world environments where explanations are required. However, it is known that the interpretability of concepts output by SENN is reduced in general settings, such as autonomous driving scenarios. Thus, this study combines contrastive learning with concept learning to improve the readability of concepts and the accuracy of tasks. We call this model {\it Contrastive Self-Explaining Neural Network} ({\it C-SENN}).
\end{abstract}

\section{Introduction}
\label{sec:introduction}
\begin{figure}[h]
\centering
\begin{minipage}{0.465\hsize}
\begin{center}
\includegraphics[width=1.0\linewidth]{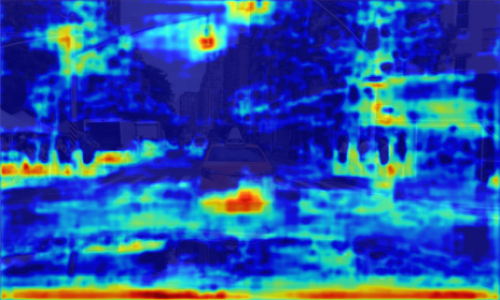}\\
(A)
\end{center}
\end{minipage}
\begin{minipage}{0.465\hsize}
\begin{center}
\includegraphics[width=1.0\linewidth]{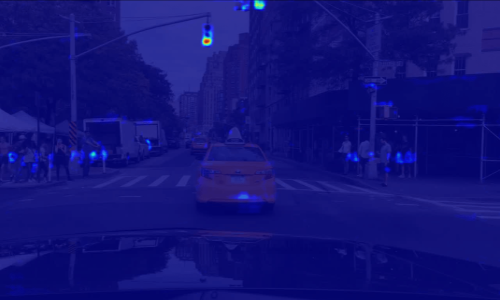}\\
(B)
\end{center}
\end{minipage}
\caption{Saliency map of concepts acquired by (A) the modified self-explaining neural network (M-SENN)~\cite{sawada2022concept_ieee} and (B) Contrastive SENN (C-SENN). The proposed method successfully acquired the concept of responding to a traffic signal. See Fig.~\ref{fig:masking} (A) for the input image.}
\label{fig:senn_c-senn_exam}
\end{figure}

When people are given an input signal, they use concepts in their brains to understand its meaning. Concepts are abstract and verbalizable information arrangements~\cite{conceptwiki}, and the construction of models, which incorporate them, is essential for model utilization in a real environment where explainability and interpretability are required. To make deep learning models more accessible to the world outside the laboratory, we build a model in which individual neurons correspond to individual concepts on a one-to-one basis.

To develop models that incorporate concepts, this study focuses on the self-explaining neural network (SENN)~\cite{alvarez2018towards,sawada2022concept_ieee}, which is a generalized linear model that acquires unsupervised concepts. Alvarez-Melis and Jaakkola~\cite{alvarez2018towards} confirmed the SENN’s ability to acquire concepts for the final output~\cite{alvarez2018towards}. However, their validation work was performed only on relatively simple image datasets, such as MNIST~\cite{lecun1998gradient} and CIFAR-10~\cite{krizhevsky2009learning}; thus, no validation was performed on datasets that capture more complex and near-real-world environments, such as Berkeley DeepDrive (BDD)~\cite{yu2020bdd100k}. Sawada and Nakamura~\cite{sawada2022concept_ieee} tackled this problem by using the discriminator and shared intermediate network (They called {\it modified SENN} ({\it M-SENN})). However, the interpretability of the concepts output by M-SENN is still low in such environments, as shown in Fig.~\ref{fig:senn_c-senn_exam} (A) (see Sec.~\ref{sec:experiments} for details).

\begin{figure}[t]
\centering
\includegraphics[width=1.0\linewidth]{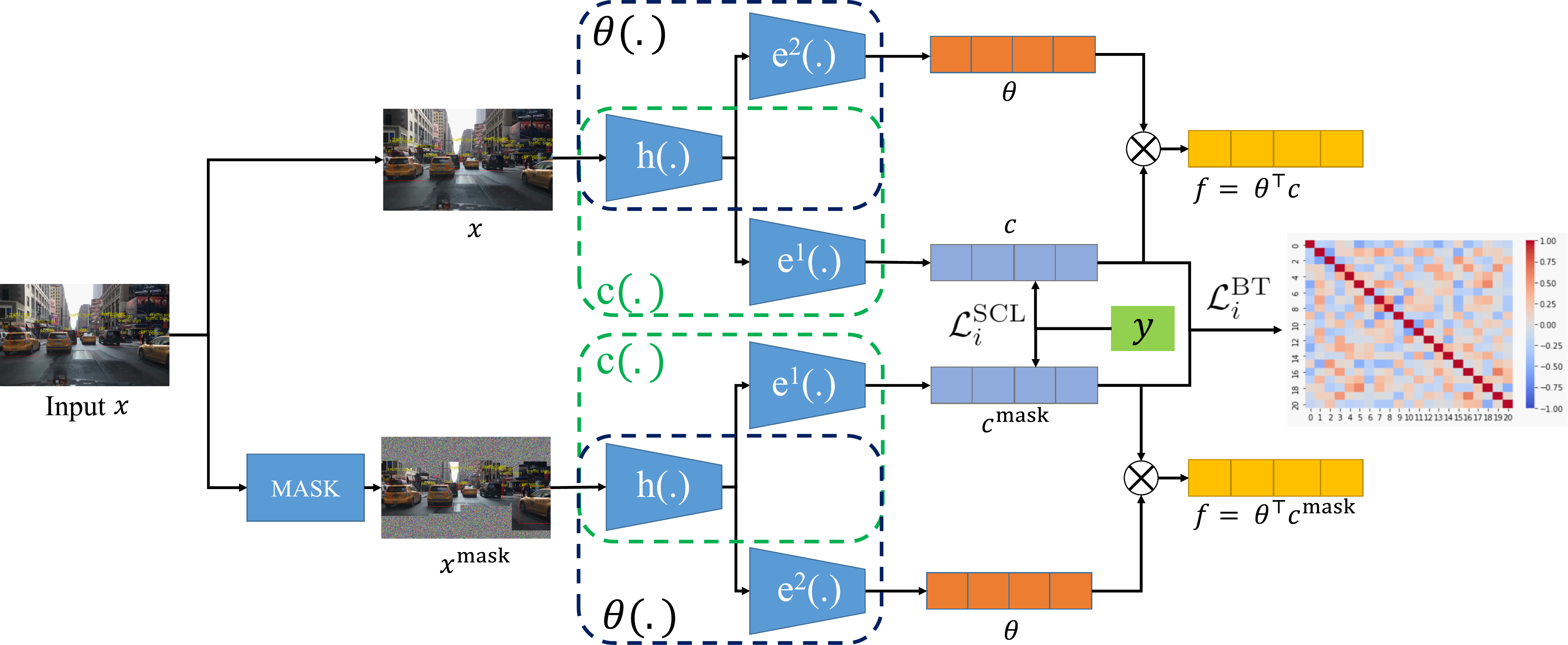}\\
\caption{Schematic diagram of the contrastive SENN (C-SENN). $\vect{e}^1(.)$ and $\vect{e}^2(.)$ represents the network after the feature extractor $\vect{h}(.)$ (since these are not important, we omit their detailed explanation here). The other notation is explained in Sec.~\ref{sub:c-senn}.}
\label{fig:network}
\end{figure}

In this study, we adopt contrastive learning~\cite{chen2020simple} to acquire concepts that are highly interpretable, even for images captured in complex environments. Contrastive learning is a self-supervised learning method that has been actively studied in recent years~\cite{chen2020simple,khosla2020supervised,zbontar2021barlow,wang2021understanding}. We utilize supervised contrastive learning (SCL)~\cite{khosla2020supervised} and the Barlow Twins (BT)~\cite{zbontar2021barlow}. SCL utilizes target labels for the final output, whereas BT ensures the independence of each feature (neuron) acquired via self-supervised learning. By incorporating these methods, we successfully acquire different concepts related to the final output without duplication. Furthermore, by augmenting contrastive learning data with the results of the Faster RCNN~\cite{frcnn}, which we use as a feature extractor, we generate concepts from verbalizable object regions. We refer to our proposed method as the {\it Contrastive SENN}({\it C-SENN}, See Fig.~\ref{fig:network}). We apply C-SENN to the BDD-OIA dataset~\cite{xu2020explainable}, finding that it improves the interpretability of the generated concepts (see Fig.~\ref{fig:senn_c-senn_exam} (B)). We also successfully achieved a higher accuracy than other methods, including conventional SENN.

\section{Conventional Methods}

\subsection{Self-Explaining Neural Network (SENN)}
\label{sub:senn}
SENN generates concepts in the layer just prior to the output layer. In its generalized form, it is expressed as the linear model, $f(\vect{x})=\vect{\theta}^\top \vect{x}$, as follows~\cite{alvarez2018towards}: 
\begin{eqnarray}
f(\vect{x}) = \vect{\theta}(\vect{x})^\top \vect{c}(\vect{x}),
\end{eqnarray}
where $\vect{x} \in \mathbb{R}^{D}$ is the input image, $\vect{\theta}(.)$ is the model that outputs weights for each concept, and $\vect{c}(.)$ represents the model (encoder) that outputs concepts. Additionally, $\vect{c}(\vect{x}) \in \mathbb{R}^{D_c}$ and $\vect{\theta}(\vect{x}) \in \mathbb{R}^{D_c \times k}$, where $D_c$ denotes the number of concepts set in advance and $k$ denotes the number of category of final output. SENN expresses $\vect{c}(.)$ and $\vect{\theta}(.)$ as neural networks optimized by the training dataset, $\{\vect{x}_i, \vect{y}_i\}_{i=1}^N$ ($\vect{y}_i$ is the $k$-dimensional vector representing the final output label) and minimizes the following loss function.
\begin{equation}
\mathcal{L}_{\rm{senn}} = \sum_i^N \mathcal{L}_y(f(\vect{x}_i), \vect{y}_i) +  \alpha \mathcal{L}_{x}(\vect{x}_i,\vect{g}(\vect{c}(\vect{x}_{i}))) + \beta \mathcal{L}_{\theta}(f(\vect{x}_i)), 
\end{equation}
where $\mathcal{L}_y$ represents the classification loss (e.g., binary-cross entropy), and $\mathcal{L}_{x}$ represents the reconstruction error. $\vect{g}(\vect{c}(\vect{x}_{i}))$ is the reconstructed image using the decoder $\vect{g}(.)$, and $\alpha$ and $\beta$ are the hyperparameters. $\mathcal{L}_{\theta}$ is the regularization term representing the stability of $\vect{\theta}(\vect{x})$ by the gradient. 
\begin{equation}
\mathcal{L}_{\theta}(f(\vect{x})) = || \vect{\nabla}_x f(\vect{x}) - ( \vect{\theta}(\vect{x})^\top \vect{J}^c_x )^\top || \approx 0,
\label{eqn:grad_senn}
\end{equation}
where $\vect{\nabla}_x f(\vect{x}) \in \mathbb{R}^{D \times k}$ is the derivative of $f(\vect{x})$, and $\vect{J}^c_x \in \mathbb{R}^{D_c \times D}$ is the Jacobian of the concept $\vect{c}(\vect{x})$ with respect to $\vect{x}$. This regularization term makes $\vect{\theta}(\vect{x})$ robust to small changes in concepts~\cite{alvarez2018towards}.

\subsection{Modified Self-Explaining Neural Network (M-SENN)}
\label{sub:m-senn}
As desribed in Sec.~\ref{sub:senn}, original SENN is the encoder-decoder model that requires a large number of parameters~\cite{hafner2019dream}. In addition, SENN must compute the Jacobian $\vect{J}^c_x \in \mathbb{R}^{D \times D}$ with respect to each input $\vect{x} \in \mathbb{R}^D$. Therefore, SENN cannot apply datasets larger than MNIST and CIFAR-10~\cite{koh2020concept}. Sawada and Nakamura~\cite{sawada2022concept_ieee} tackled this problem by using the discriminator and shared intermediate network as follows.
\begin{eqnarray}
\mathcal{L}_{{\rm SENN}} &=&\sum_i^N \big(  \mathcal{L}_y(f(\vect{x}_i),\vect{y}_i) + \alpha \mathcal{L}^{{\rm DIS}}(\vect{z}_i,\vect{z}_i^\prime) +\beta \mathcal{L}_{\theta}(f(\vect{x}_i)) \big),
\label{eqn:senn}
\end{eqnarray}
where $\vect{z} = [\vect{c}^\top, \vect{h}(\vect{x})]^\top$, $\vect{z}^\prime = [\vect{c}^\top, \vect{h}(\vect{x}^\prime)]^\top$, and $\vect{x}^\prime$ represents an image different from $\vect{x}$. $\vect{h}(.)$ ($\vect{h}(\vect{x}) \in \mathbb{R}^{D_h}$) is the intermediate network. We use the output of backbone with FPN of Faster RCNN ~\cite{frcnn}. $\mathcal{L}^{{\rm DIS}}$ represents the squared error~\cite{sawada2022concept_ieee}.
\begin{equation}
\mathcal{L}^{\rm{DIS}}(d(\vect{z}),d(\vect{z}^{\prime})) = || d(\vect{z}) - a ||^2 + || d(\vect{z}^{\prime}) - b ||^2,
\label{eqn:dis}
\end{equation}
where $a = 1$, $b = 0$, and $d(.)$ represent the discriminator. Since the number of parameters of $d(.)$ is much smaller than the decoder's one, \cite{sawada2022concept_ieee} used this discriminator loss $\mathcal{L}^{\rm{DIS}}$ instead of the reconstruction error $\mathcal{L}_{x}$. Additionally, $\mathcal{L}_{\theta}$ represents a constraint term that constrains the variation of $\vect{\theta}(.)$ as follows:
\begin{equation}
\mathcal{L}_{\theta}(f(\vect{x})) = || \vect{\nabla}_h f(\vect{x}) - ( \vect{\theta}(\vect{x})^\top \vect{J}^{c}_h )^\top || \approx 0,
\label{eqn:grad_senn}
\end{equation}
where $\vect{\nabla}_h f(\vect{x}) \in \mathbb{R}^{D_h \times k}$ is the derivative of $f(\vect{x})$ with respect to the intermediate feature,  $\vect{J}^{c}_h$ is the Jacobian of $D_c \times D_h$, and the relationship between $\vect{\nabla}_x f(\vect{x})$ and $\vect{\nabla}_h f(\vect{x})$, and 
$\vect{J}^{c}_h$ and $\vect{J}^{c}_x$ are as follows:
\begin{eqnarray}
\vect{\nabla}_x f(\vect{x}) &=& ( \vect{\nabla}_h f(\vect{x})^\top \vect{J}^h_x )^\top,\\
\vect{J}^{c^{\rm{im}}}_x &=& \vect{J}^c_h \vect{J}^h_x.
\end{eqnarray}
By using these techniques, \cite{sawada2022concept_ieee} achieved to train models for the BDD-OIA~\cite{xu2020explainable} and CUB-200-2011~\cite{wah2011caltech}.

\section{Contrastive Self-Explaining Neural Network (C-SENN)}
\label{sub:c-senn}
As noted in Section~\ref{sec:introduction}, the interpretability of concepts generated by M-SENN is low in complex environments. One of the reasons is that it is challenging for the true/false discriminator to generate concepts while maintaining the similarity in the input space. In this study, we focus on contrastive learning to better maintain the similarity in the input space.

Contrastive learning is a kind of self-supervised learning that has been actively studied in recent years~\cite{chen2020simple,chen2021exploring}. Many contrastive learning methods combine multiple data augmentation techniques and learn to place transformations from the same image closer together in the feature space and those from different images farther apart. Then, the top layer (projection head) is removed, and the rest is used as a feature extractor. In contrast, as in \cite{fu2021scala}, we minimize the contrastive loss simultaneously with other losses. 

\begin{figure}[t]
\centering
\begin{minipage}{0.465\hsize}
\begin{center}
\includegraphics[width=1.0\linewidth]{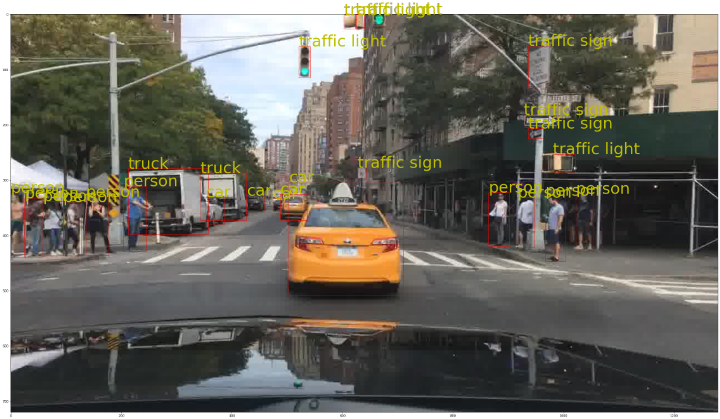}\\
(A)
\end{center}
\end{minipage}
\begin{minipage}{0.465\hsize}
\begin{center}
\includegraphics[width=1.0\linewidth]{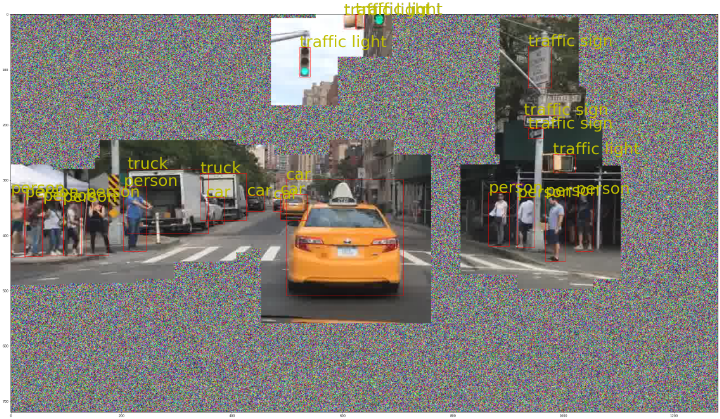}\\
(B)
\end{center}
\end{minipage}
\caption{Example of a mask image based on the output of Faster RCNN.}
\label{fig:masking}
\end{figure}

This study utilizes supervised contrastive learning (SCL)~\cite{khosla2020supervised} and the Barlow Twins (BT)~\cite{zbontar2021barlow} for generating concepts.
\begin{equation}
\mathcal{L}_{{\rm SCL}}(\vect{c}_i, \mathcal{A}_{\backslash i}) = \frac{1}{|\mathcal{P}_i|} \sum_{p \in \mathcal{P}_i} {\rm log} \frac{{\rm exp}(\vect{c}_i^\top \vect{c}_p / \tau)}{\sum_{a \in \mathcal{A}_{\backslash i}} {\rm exp}(\vect{c}_i^\top \vect{c}_a / \tau)},
\label{eqn:scl}
\end{equation}
\begin{equation}
\mathcal{L}_{{\rm BT}}(\mathcal{A}) = \sum_{j}^{D_c} (1-R_{j,j})^2 + \lambda \sum_{j}^{D_c} \sum_{k \neq j}^{D_c} R_{j,k}^2,
\label{eqn:barlow}
\end{equation}
where $\mathcal{L}^{{\rm SCL}}$ is the SCL loss function and $\mathcal{L}^{{\rm BT}}$ is the BT loss function. Furthermore,
\begin{equation}
R_{j,k} = \frac{\sum_i^N c_{j,i} c_{k,i}^{{\rm mask}}}{\sqrt{\sum_i^N (c_{j,i})^2}\sqrt{\sum_i^N (c^{\rm mask}_{k,i})^2}},
\label{eqn:barlow_R}
\end{equation}
where $\lambda$ is the hyperparameter, $\mathcal{A}_{\backslash i}$ is the set of concepts of all data except the $i$-th data, and $\mathcal{A} = \{ \vect{c}_i, \mathcal{A}_{\backslash i} \}$. $\mathcal{P}_i$ is the set of concepts in the data having the same label as the $i$-th target label, and $|\mathcal{P}_i|$ is the total number of such concepts. $c_{j,i}^{{\rm mask}}$ represents the $j$-th concept of $\vect{x}_i^{{\rm mask}}$, which is a result of data augmentation of $\vect{x}_i$. We utilize the detection results of Faster RCNN~\cite{frcnn} for data augmentation. Note that the intermediate extractor $\vect{h}(.)$ is part of this Faster RCNN pretrained by the COCO~\cite{coco} and fine-tuned with BDD100k~\cite{yu2020bdd100k}. We detect bounding boxes with Faster RCNN, and replace the regions other than the bounding box$+\epsilon$ ($\epsilon$ is a constant) with white noise. Then, we treat the resulting image as $\vect{x}_i^{{\rm mask}}$ (Fig.~\ref{fig:masking}). As a result, the model learns to acquire concepts from the periphery of the object region. Notably, $\vect{c}^{\rm{mask}}=[c_1^{\rm{mask}}, c_2^{\rm{mask}}, \cdots, c_{D_c}^{\rm{mask}}]^\top$  is included in $\mathcal{P}$.

By combining Equations (\ref{eqn:scl}) and (\ref{eqn:barlow}) with Equation (\ref{eqn:senn}), the final loss function of C-SENN is as follows:
\begin{eqnarray}
\mathcal{L} &=& \sum_i^N \big( \mathcal{L}_y(f(\vect{x}_i),\vect{y}_i) + \beta \mathcal{L}_{\theta}(f(\vect{x}_i)) +\lambda_{{\rm SCL}} \mathcal{L}_{{\rm SCL}}(\vect{c}_i, \mathcal{A}_{\backslash i}) \big) + \lambda_{{\rm BT}} \mathcal{L}_{{\rm BT}}(\mathcal{A}) .
\end{eqnarray}
where $\lambda_{{\rm SCL}}$ and $\lambda_{{\rm BT}}$ represent hyperparameters that control the loss function’s value. Notably, $\mathcal{L}_{{\rm DIS}}$ (the distance measure loss) is not used here. The purpose of $\mathcal{L}_{{\rm DIS}}$ is to preserve distance relations in the input image and concept space~\cite{sawada2022concept_ieee}. However, the same result is obtained using the SCL loss $\mathcal{L}_{{\rm SCL}}$ clearly from Equation (\ref{eqn:scl}). Thus, because we consider $\mathcal{L}_{{\rm DIS}}$ to be substitutable by $\mathcal{L}_{{\rm SCL}}$, we do not use $\mathcal{L}_{{\rm DIS}}$ to reduce redundancy. In this study, the resultant SENN is termed ``C-SENN''.

Figure~\ref{fig:network} shows a schematic diagram of C-SENN. As shown, for computational efficiency, only the Faster RCNN masking described above is used for data augmentation. Similarly, the original image is used as input without masking during inferencing.

\section{Experimental Results}
\label{sec:experiments}
This section describes our evaluation of C-SENN using the BDD-OIA~\cite{xu2020explainable} for autonomous driving, consisting of 16,802 training mages, 2,270 evaluation images, and 4,572 test images. Each image is assigned target labels (``forward ($y_{i,1}$)'', ``stop ($y_{i,2}$)'', ``right turn ($y_{i,3}$)'', and ``left turn ($y_{i,4}$)''), representing possible driving actions (namely, $k=4$). Notably, BDD-OIA has 21 concept labels (e.g., ``green light'' or ```good visibility'') for each image, but these are not used for C-SENN and M-SENN training (original SENN could not run). 

In this article, we set $\beta=0.01$, $\lambda=0.1$, $\lambda_{\rm SCL}=1.0$, $\lambda_{\rm BT}=0.001$, and $D_c=21$ to equal the number of BDD-OIA concepts. The network structure and other hyperparameters are the same as those in \cite{sawada2022concept_ieee}, and all experiments were performed using one NVIDIA Tesla v100 GPU with 32-GB memory.

\begin{figure}[t]
\centering
\includegraphics[width=0.45\linewidth]{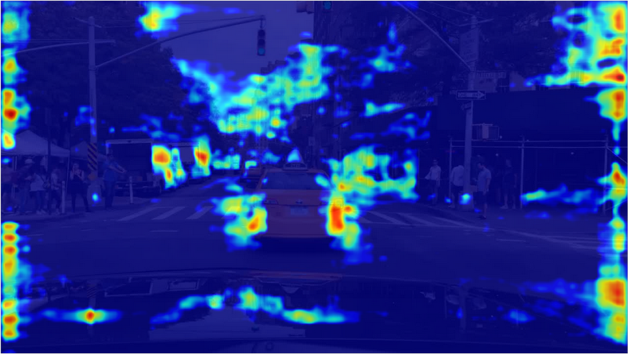}
\includegraphics[width=0.45\linewidth]{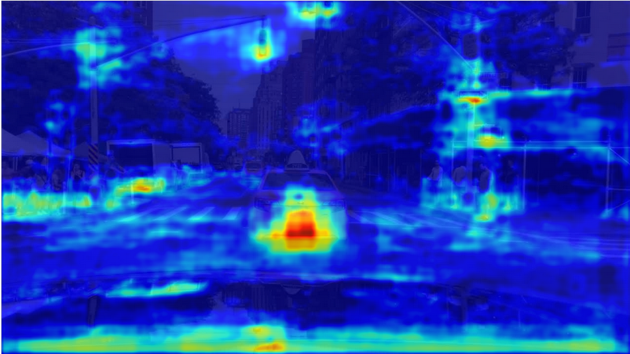}\\
SENN\\
\includegraphics[width=0.45\linewidth]{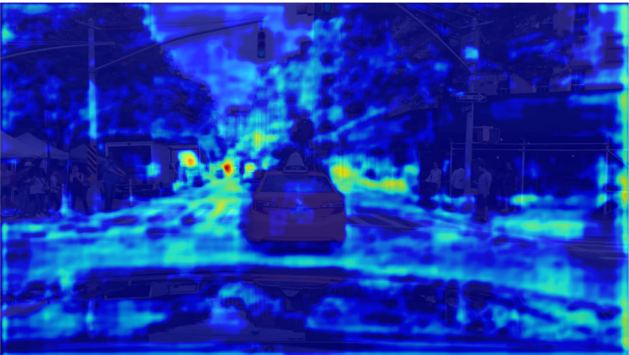}
\includegraphics[width=0.45\linewidth]{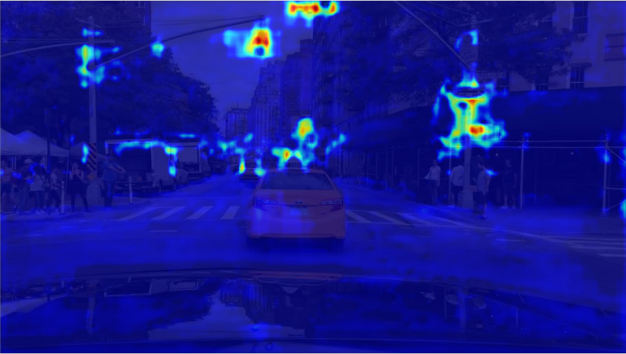}\\
SC-SENN\\
\includegraphics[width=0.45\linewidth]{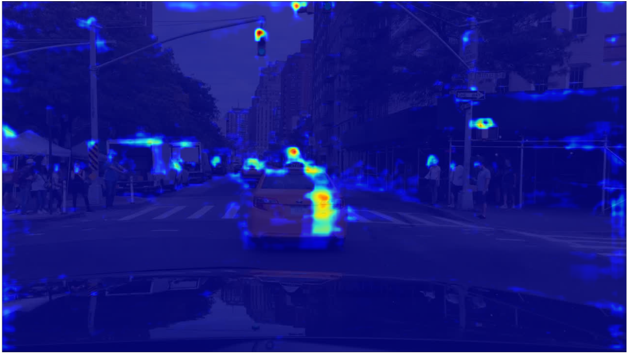}
\includegraphics[width=0.45\linewidth]{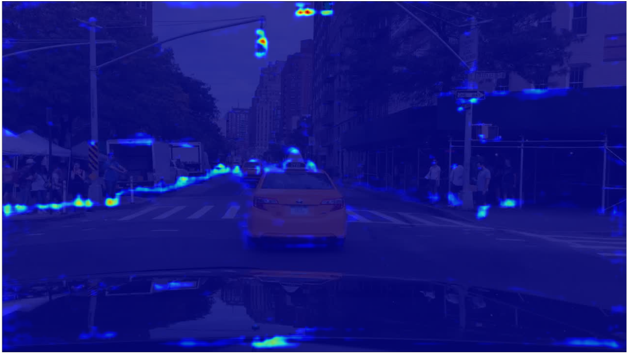}\\
C-SENN
\caption{Examples of concept saliency maps obtained by each method. See Fig.~\ref{fig:masking} (A) for the input image. The supervised contrastive (SC) self-explaining neural network (SENN) model replaces the $\mathcal{L}_{{\rm DIS}}$ of the original SENN with $\mathcal{L}_{{\rm SCL}}$ and does not use the BT loss function.}
\label{fig:gradcam_comp}
\end{figure}

Figure~\ref{fig:gradcam_comp} shows examples of concept saliency maps obtained by each method. These maps were generated by the grad-CAM~\cite{selvaraju2017grad}. Notably, the SC-SENN model replaced the $\mathcal{L}_{{\rm DIS}}$ of the SENN with $\mathcal{L}_{{\rm SCL}}$ and did not use the BT loss function. With $\mathcal{L}_{{\rm SCL}}$, each concept was focused on a more local region in the image. For example, the SC-SENN image on the right side of Fig.~\ref{fig:gradcam_comp} focuses on traffic signals and signs. The addition of $\mathcal{L}_{{\rm BT}}$ improved the focus to even finer regions. For example, the C-SENN image on the left side of Fig.~\ref{fig:gradcam_comp} focuses on the right side of each object, whereas the image on the right focuses on the foot of each. The former is a useful concept for the ``right turn'' action, whereas the latter is a useful concept for ``forward'' and ``stop'' actions. The results of the C-SENN saliency map for other images are shown in the appendix.

\begin{figure}[t]
\centering
\begin{minipage}{0.465\hsize}
\begin{center}
\includegraphics[width=1.0\linewidth]{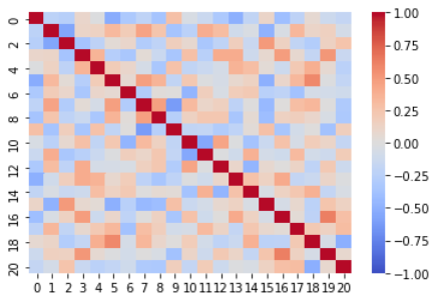}\\
(A)
\end{center}
\end{minipage}
\begin{minipage}{0.465\hsize}
\begin{center}
\includegraphics[width=1.0\linewidth]{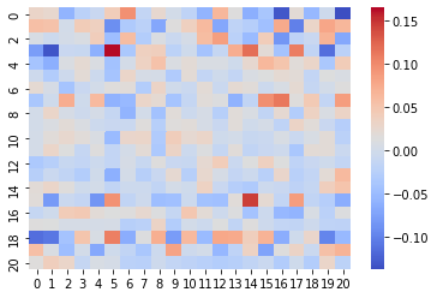}\\
(B)
\end{center}
\end{minipage}
\caption{(A): Correlation matrix between concepts acquired by the C-SENN; (B): Correlation matrix between each concept and the concept label.}
\label{fig:corr}
\end{figure}

Figure~\ref{fig:corr} (A) shows the correlation matrix between concepts acquired by C-SENN, and Fig.~\ref{fig:corr} (B) shows the correlation matrix of C-SENN concepts and concept labels in the BDD-OIA dataset. Fig.~\ref{fig:corr} (A) shows that the correlation between concepts was small; thus, multiple concepts can be acquired by C-SENN without duplication. This is the effect of SCL. In contrast, Fig.~\ref{fig:corr} (B) shows that the concepts acquired by C-SENN had little correlation with the concept labels. Thus, the C-SENN makes predictions based on concepts that differ from those deemed necessary by human annotators. For example, concepts corresponding to the foot and right sides of each object, as shown in the C-SENN image in Fig.~\ref{fig:gradcam_comp}, do not exist in the teacher concepts. This suggests that it is possible to discover new insights by checking the concepts generated by C-SENN. Additional analyses are shown in the appendix.

\begin{figure}[t]
\centering
\begin{minipage}{0.465\hsize}
\includegraphics[width=0.835\linewidth]{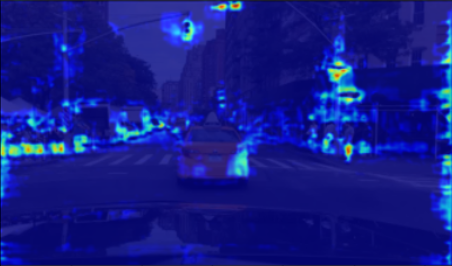}\\
\includegraphics[width=1.0\linewidth]{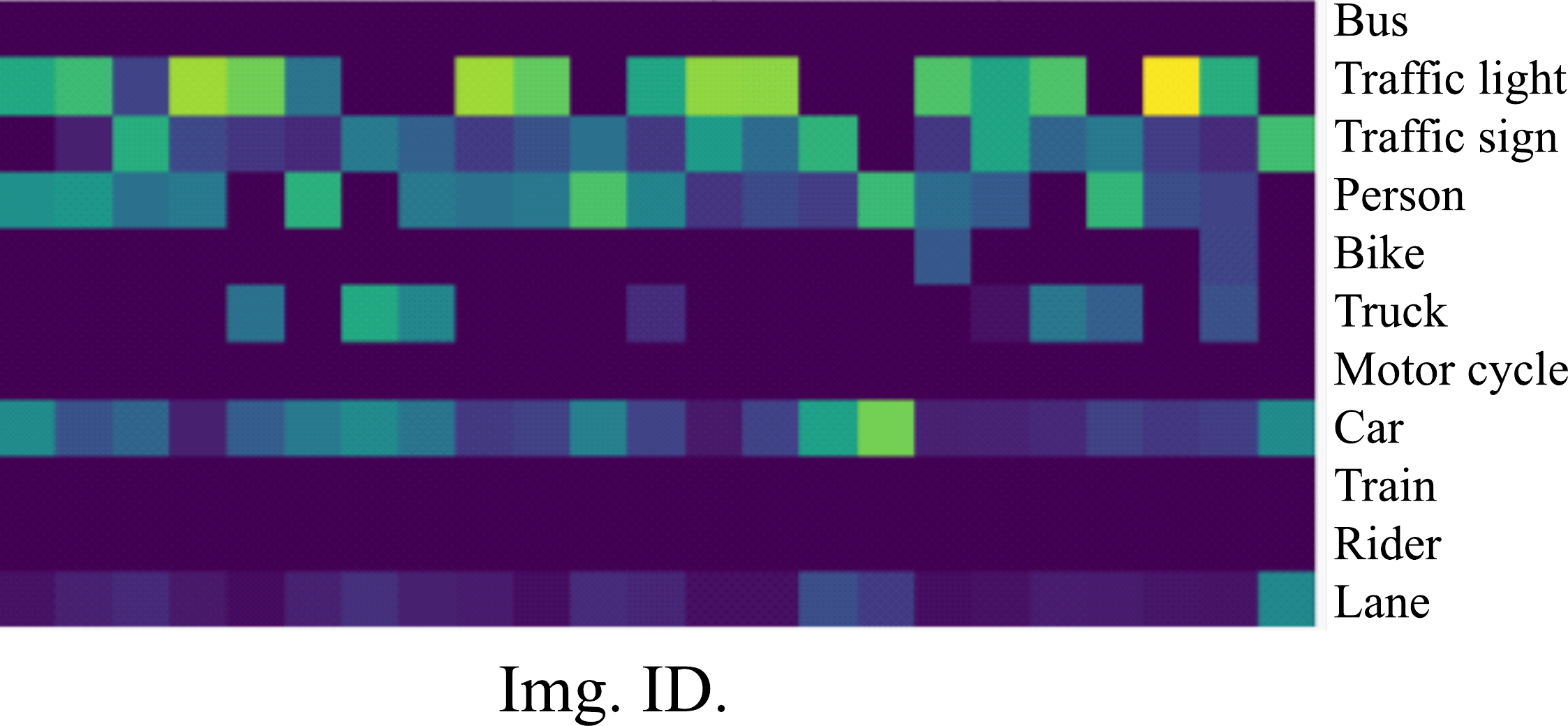}\\
\end{minipage}
\begin{minipage}{0.465\hsize}
\includegraphics[width=0.835\linewidth]{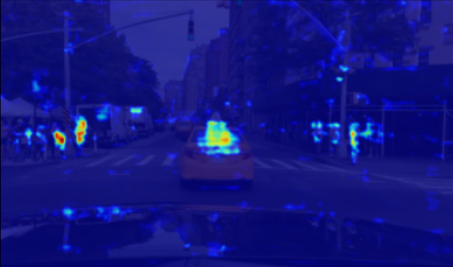}\\
\includegraphics[width=1.0\linewidth]{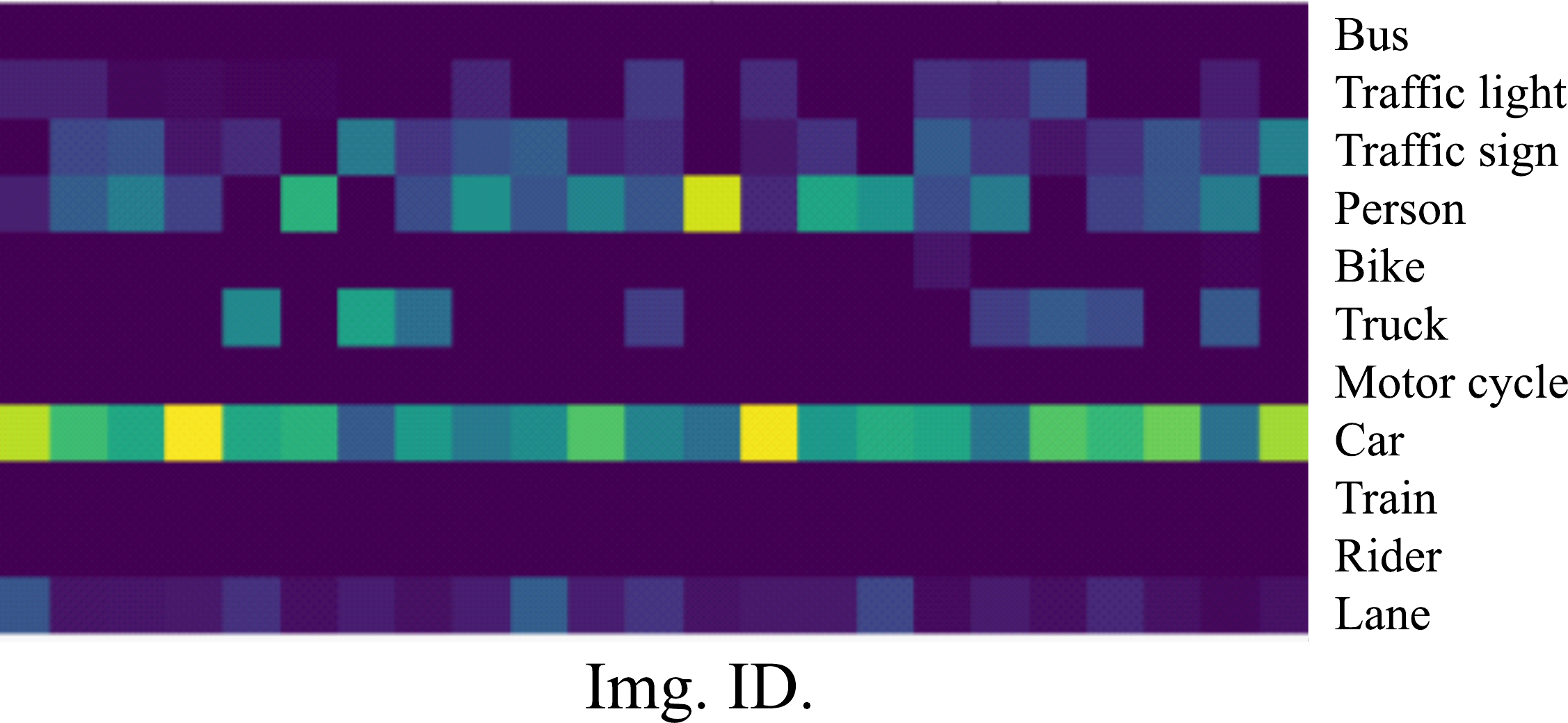}\\
\end{minipage}
\caption{Examples of regions in each image where each concept is focused. Top: Concept saliency map (input image is Fig.~\ref{fig:masking} (A)); Bottom: Regions of each input image where each concept is focused. X-axis: Image ID; Y-axis: Objects detected by the Faster RCNN. The brighter the color, the more attention is paid.}
\label{fig:gradcam_matrix}
\end{figure}

Next, Fig.~\ref{fig:gradcam_matrix} shows examples of the regions in each image where each concept was focused. In the bottom part of the figure, the X-axis represents image ID, and the Y-axis represents the object detected by Faster RCNN. The higher the average gradCAM score in each object region, the brighter the display. This confirms that C-SENN generates concepts that span multiple object regions. Concerning interpretability, it is desirable that concepts be generated from a smaller number of objects. This is a topic for future research.

\begin{table}[t]
\centering
{\begin{tabular}{c|cccc|c}
\hline
Model & F & S & R & L & $mF1$ \\
\hline \hline
Vanilla & 0.54 & 0.666 & 0.11 & 0.151 & 0.367 \\
CBM~\cite{koh2020concept} & 0.795 & 0.732 & 0.431 & 0.483 & 0.610 \\
M-SENN~\cite{sawada2022concept_ieee} & 0.705 & 0.727 & 0.339 & 0.385 & 0.539 \\
C-SENN & 0.772 & 0.744 & 0.486 & 0.469 & 0.618 \\
\hline
\end{tabular}}\\
\caption{Comparison of identification accuracy. F, S, R, and L are the final output labels in the BDD-OIA dataset corresponding to ``forward,'' ``stop,'' ``turn right,'' and ``turn left,'' respectively. $mF1$ represents the average of the F1 score for each action.}
\label{tbl:BDD}
\end{table}

Finally, we show the driving behavior recognition results using each method.
Table~\ref{tbl:BDD} shows result. Note that vanilla is the model that does not explicitly generate concepts between input and output, and the concept bottleneck model (CBM)~\cite{koh2020concept} is the model that trains by utilizing concept labels assigned in the BDD-OIA dataset. The evaluation indices used were the F1-score for each driving behavior and $mF1$, which is the average F1-score for each action~\cite{xu2020explainable}. Comparing the results to vanilla confirms that recognition accuracy improved by generating concepts between input and output. Additionally, the results show that C-SENN was more accurate than M-SENN. C-SENN also showed higher values than CBM in many indices. This suggests that C-SENN can generate more appropriate recognition concepts than human-annotator preassigned concept labels. Although CBM offers the potential to improve accuracy by refining the concept labels related to driving behaviors in advance, it makes the annotation costs enormous.

\section{Conclusion}
In this study, we proposed C-SENN, which combines the M-SENN with two types of contrastive learning, BT and SCL. We conducted experiments on the BDD-OIA dataset, confirming that C-SENN acquired concepts that were more highly interpretable than those of other methods, and it recognizes driving actions with high accuracy. In the future, we plan to conduct a detailed analysis of what constitutes a highly interpretable concept through functional evaluations. However, collaboration with various researchers, such as psychologists, will be essential to know the ideal form of the concepts we are acquiring. This is not low-hanging fruit, but it is necessary to work on over the long term.

\bibliographystyle{unsrt}
\small{\bibliography{main.bib}}

\section*{Appendix}
\setcounter{section}{0}
\setcounter{table}{0}
\setcounter{figure}{0}
\def\thesection{\Alph{section}}
\def\thetable{\Alph{table}}
\def\thefigure{\Alph{figure}}

\section{Saliency Maps for Other Images}
\begin{figure}[h]
\centering
\includegraphics[width=0.925\linewidth]{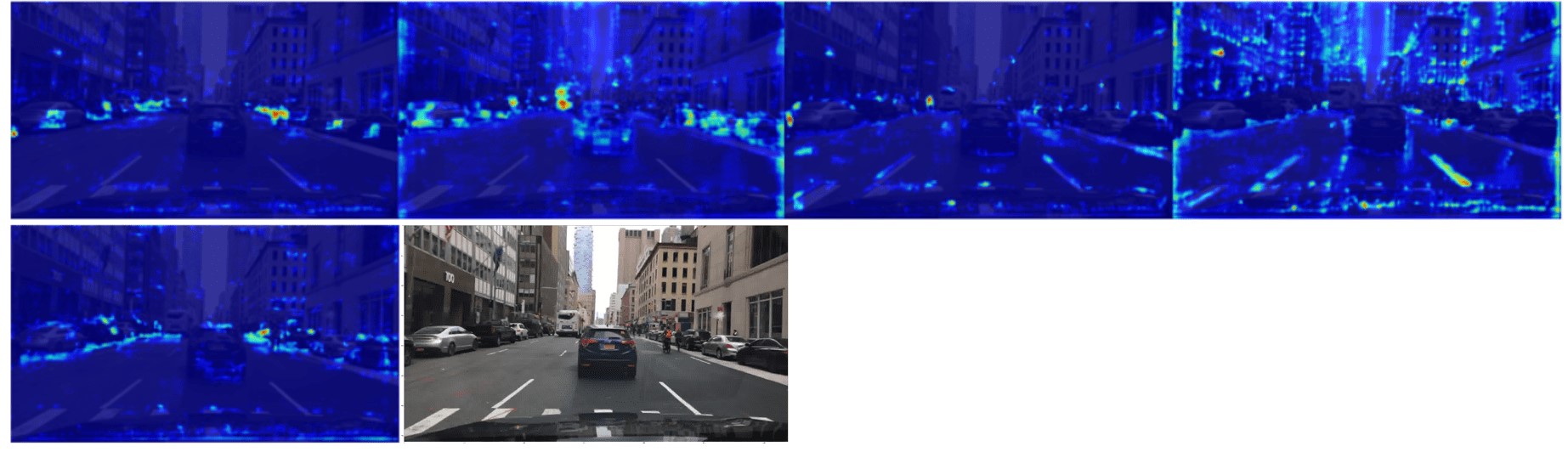}\\
\includegraphics[width=0.925\linewidth]{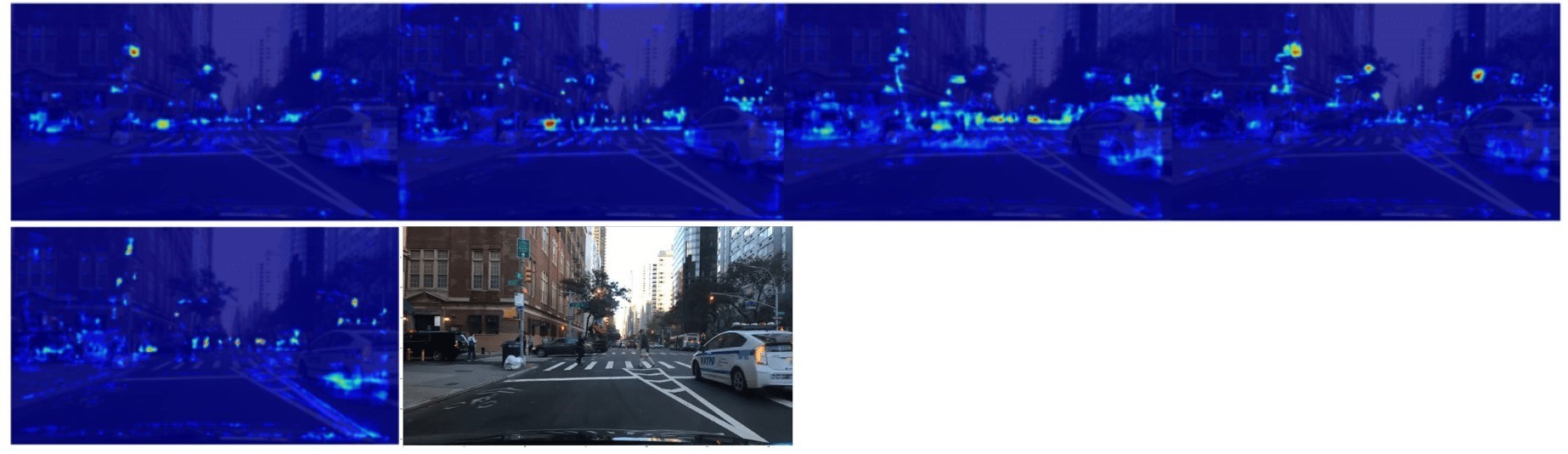}\\
\includegraphics[width=0.925\linewidth]{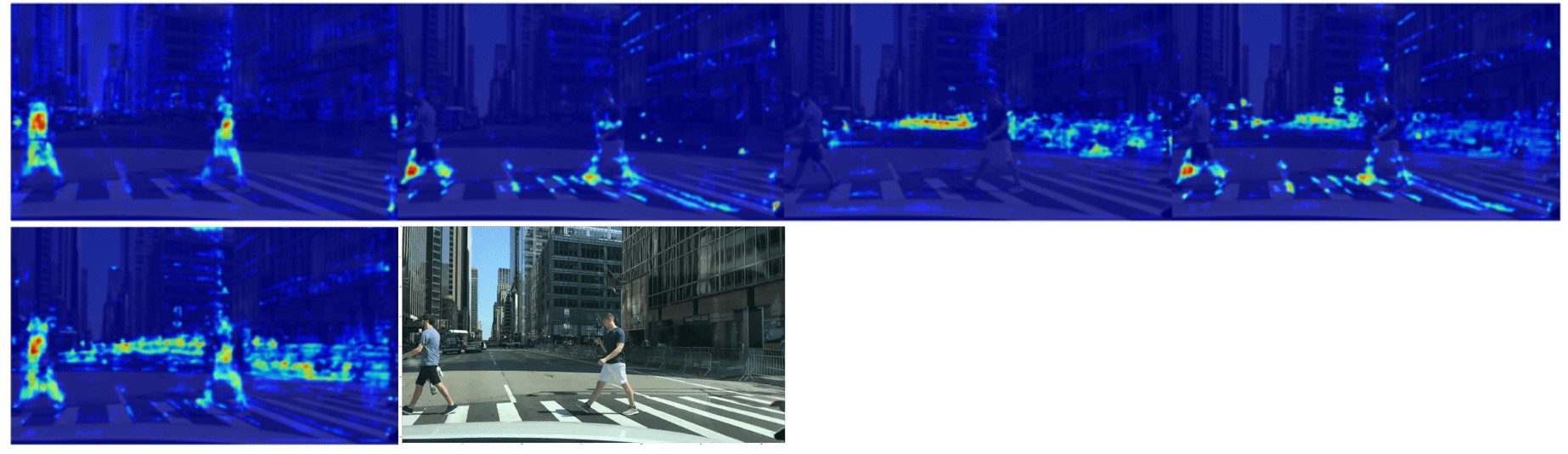}\\
\includegraphics[width=0.925\linewidth]{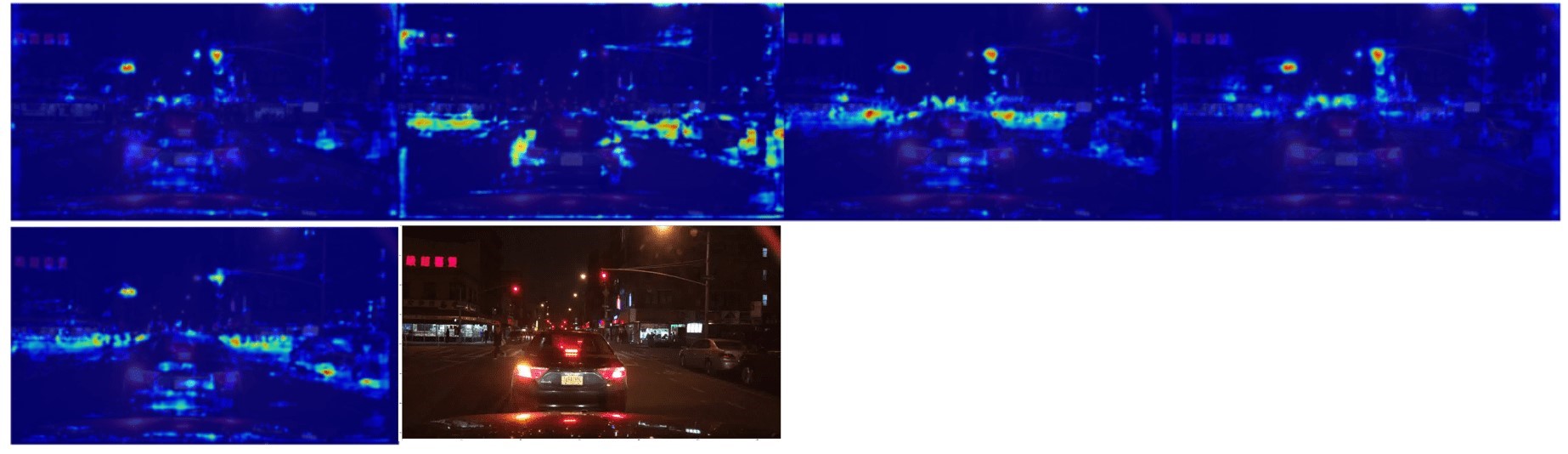}\\
\includegraphics[width=0.925\linewidth]{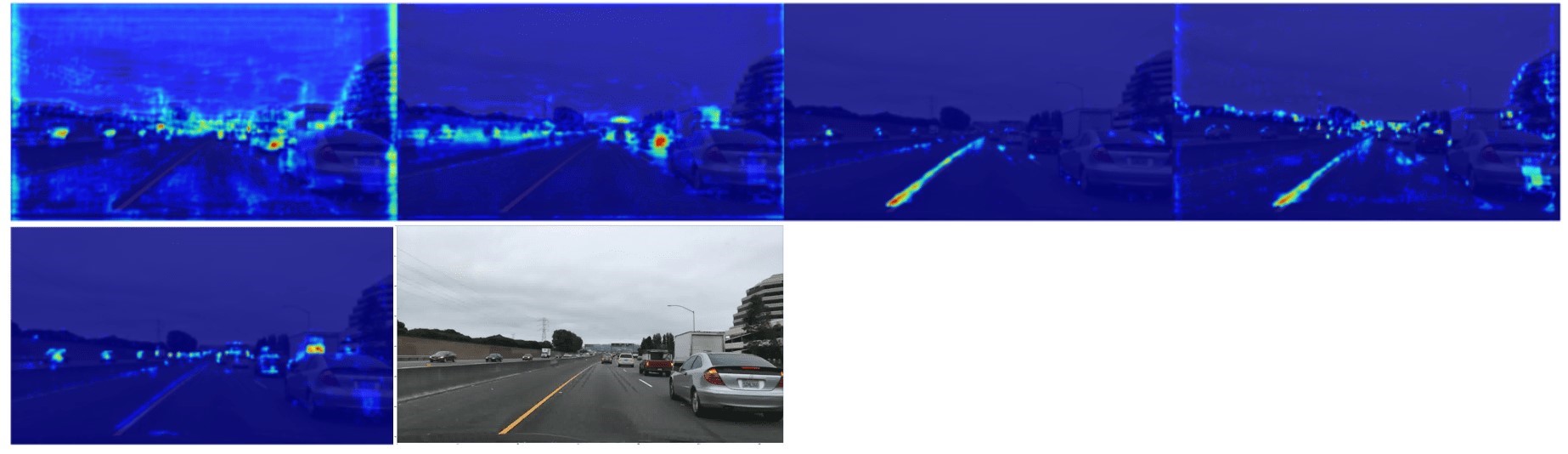}\\
\includegraphics[width=0.925\linewidth]{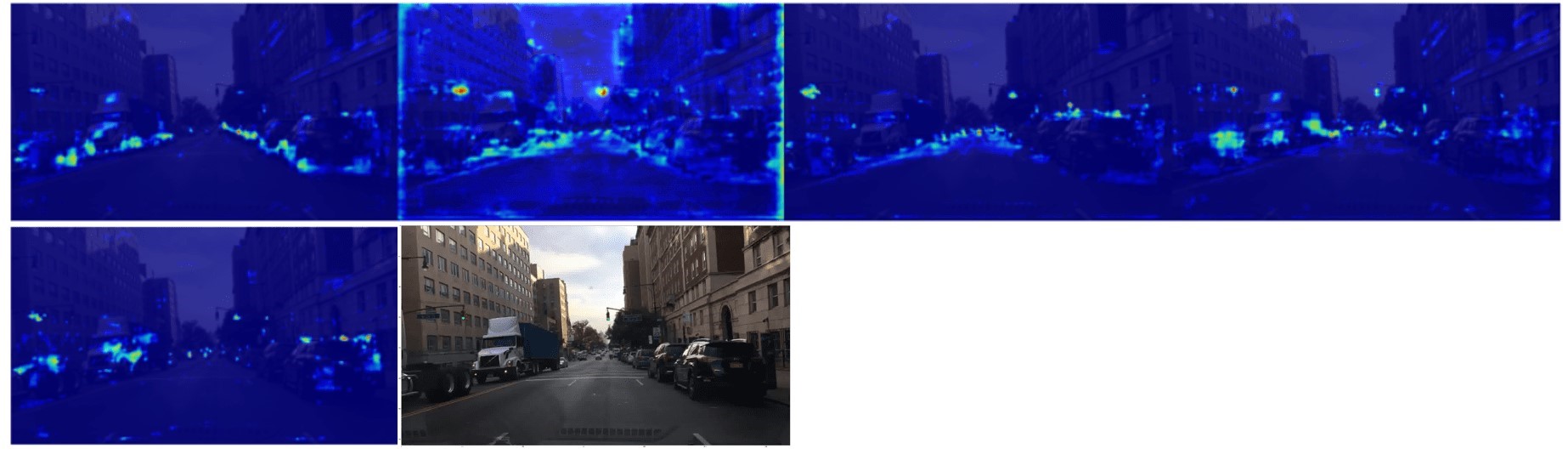}\\
\caption{Saliency maps for other images except Fig.~\ref{fig:gradcam_comp} obtained by C-SENN.}
\label{fig:gradcam_others}
\end{figure}

Figure~\ref{fig:gradcam_others} shows the C-SENN saliency maps for other images. Bottom right-most images are input, and other images are saliency maps obtained by the grad-CAM. As shown in these figures, some neurons focus on small areas, while others focus on the whole image with few obstacles and good visibility. When driving good visibility environments, humans also focus on the entire field of view. Therefore, we consider that neurons concentrating over the whole image corresponds to this human action.

\section{Additional Analysis}
Here, we show some results when changing conditions. 

Figure~\ref{fig:corr} (A) and (B) show the correlation matrix using Barlow Twins~\cite{zbontar2021barlow} (same as Fig.~\ref{fig:gradcam_matrix}(A)) and Variance-Invariance-Covariance Regularization (VICReg)~\cite{bardes2021vicreg}, respectively. VICReg is also a method that ensures the independence of each concept, as in BT. Comparing these matrices, BT has smaller values of the non-diagonal components than VICReg. Therefore, we used the BT for C-SENN.

\begin{figure}[h]
\centering
\includegraphics[width=1.0\linewidth]{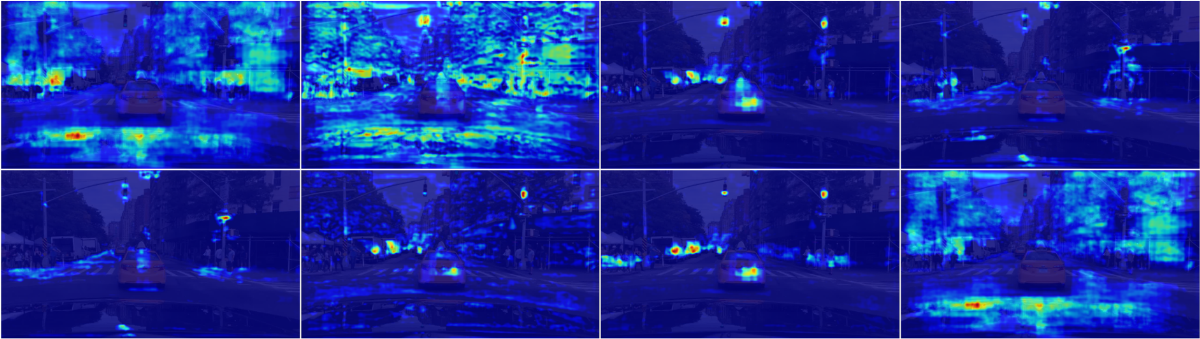}\\
$D_c=10$\\
\includegraphics[width=1.0\linewidth]{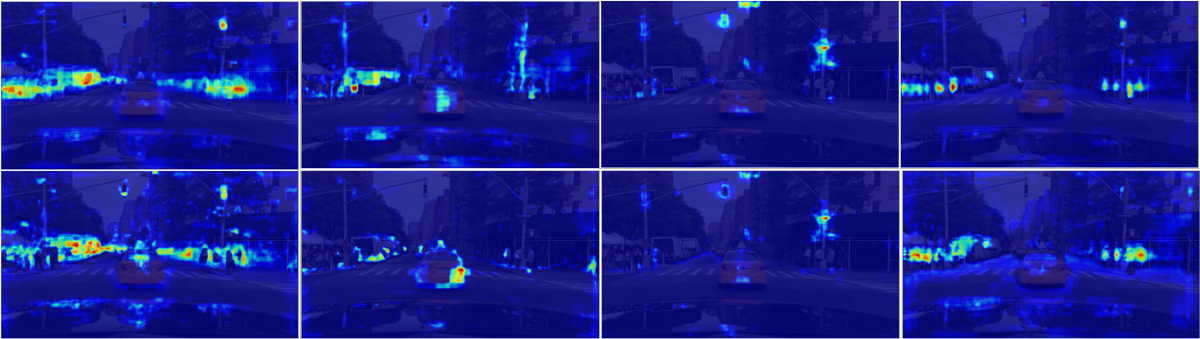}\\
$D_c=21$\\
\caption{Saliency maps when changing the number of concepts $D_c$. Input image is Fig.~\ref{fig:masking} (A).}
\label{fig:d_c}
\end{figure}

Figure~\ref{fig:corr} (C) and (D) show the correlation matrix when changing the number of concepts $D_c$. These figures show that C-SENN has smaller values of the non-diagonal components irrelevant to $D_c$. Figure~\ref{fig:d_c} shows the saliency maps of $D_c=10$ and $D_c=21$. When increasing $D_c$, C-SENN focuses on small areas and vice-vasa. Namely, generated concepts change by changing $D_c$. Automatic acquisition of the appropriate number of concepts, taking interpretability and recognition rate into account, is an issue for future work.

\begin{figure}[t]
\centering
\begin{minipage}{0.465\hsize}
\begin{center}
\includegraphics[width=1.0\linewidth]{fig/corr.png}\\
(A)
\end{center}
\end{minipage}
\begin{minipage}{0.465\hsize}
\begin{center}
\includegraphics[width=1.0\linewidth]{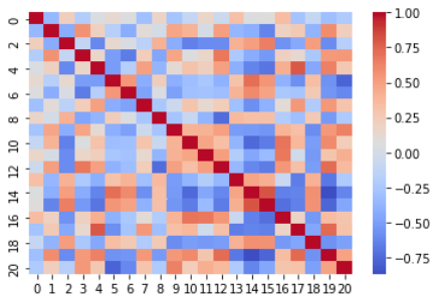}\\
(B)
\end{center}
\end{minipage}
\begin{minipage}{0.465\hsize}
\begin{center}
\includegraphics[width=1.0\linewidth]{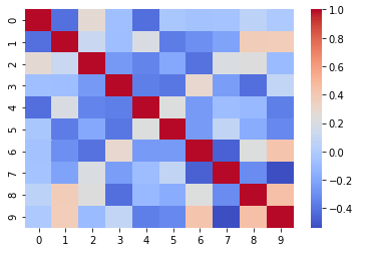}\\
(C)
\end{center}
\end{minipage}
\begin{minipage}{0.465\hsize}
\begin{center}
\includegraphics[width=1.0\linewidth]{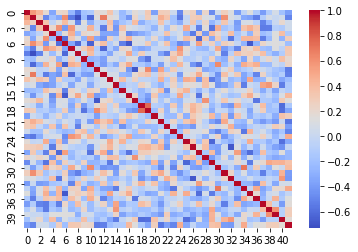}\\
(D)
\end{center}
\end{minipage}
\caption{Correlation matrices using (A) BT~\cite{zbontar2021barlow} and (B) VICReg~\cite{bardes2021vicreg}. (C) and (D) are correlation matrices when changing $D_c=10$ and $D_c=41$, respectively. Note that these are generated by BT.}
\label{fig:corr}
\end{figure}

\end{document}